\documentclass[runningheads]{llncs}

\usepackage{graphicx}
\usepackage{comment}
\usepackage{amsmath,amssymb}
\usepackage{color}
\usepackage{url}
\usepackage{hyperref}
\usepackage{tikz}
\usepackage{comment}
\usepackage{array}
\usepackage{xcolor}
\usepackage{cite}

\newcolumntype{W}{>{\centering\arraybackslash}p{0.3\textwidth}}
\newcolumntype{C}{>{\centering\arraybackslash}p{0.25\textwidth}}
\newcolumntype{B}{>{\centering\arraybackslash}p{0.2\textwidth}}
\newcolumntype{A}{>{\centering\arraybackslash}p{0.15\textwidth}}

\newcommand{\etal}{\textit{et al.}}
\newcommand{\ie}{\textit{i.e.}}

\newif\ifreview
\reviewfalse

\ifreview
	\usepackage{lineno}

	\linenumbers
\fi

\begin{document}


\def\SubNumber{013}

\def\GCPRTrack{Main Track}

\title{Self-supervised Learning for Unintentional Action Prediction}

\ifreview
\else

	\author{Olga Zatsarynna\inst{1} \and
	Yazan Abu Farha\inst{2} \and
	Juergen Gall\inst{1}}
	
	\authorrunning{Zatsarynna et al.}
	
	\institute{University of Bonn \and Birzeit University \\
	\email{\{zatsarynna, gall\}@iai.uni-bonn.de} \\
	\email{yabufarha@birzeit.edu}}
\fi

\maketitle              

\begin{abstract}
Distinguishing if an action is performed as intended or if an intended action fails is an important skill that not only humans have, but that is also important for intelligent systems that operate in human environments. Recognizing if an action is unintentional or anticipating if an action will fail, however, is not straight-forward due to lack of annotated data. While videos of unintentional or failed actions can be found in the Internet in abundance, high annotation costs are a major bottleneck for learning networks for these tasks. In this work, we thus study the problem of self-supervised representation learning for unintentional action prediction. While previous works learn the representation based on a local temporal neighborhood, we show that the global context of a video is needed to learn a good representation for the three downstream tasks: unintentional action classification, localization and anticipation. In the supplementary material, we show that the learned representation can be used for detecting anomalies in videos as well.

\end{abstract}
%
%
%

\section{Introduction}
Recognition of human intentions is a task that we perform easily on a daily basis. We can tell the difference between the deliberately and accidentally dropped object; between the liquid poured with intent or spilled out of carelessness. Moreover, we can easily identify the point at which a certain action turns from being intentional to an unintentional one. Our ability to do so signifies the importance of this skill in the day to day life since the intent of a particular event dictates our reaction to it. Therefore, if intelligent agents are to exist among human beings, they should possess the skill to differentiate between intentional and unintentional actions.

While there are large amounts of videos, which contain unintentional actions, freely available on various video platforms, they are not annotated. Since the videos contain an action that becomes unintentional over time, it would be very time-consuming to annotate the transition between an intentional and unintentional action in each video. To mitigate the effort required for annotation, Epstein~\etal\cite{Epstein_2020_CVPR} proposed to consider unintentional action prediction from the unsupervised learning perspective. They collected a dataset of fail videos (Oops dataset~\cite{Epstein_2020_CVPR}) from Youtube and proposed three methods for learning unintentional video features in a self-supervised way: Video Speed, Video Sorting and Video Context. Video Speed learns features by predicting the speed of videos sampled at 4 different frame rates. Video Sorting learns representations by predicting permutations of 3 clips sampled with a 0.5 second gap. Finally, Video Context relies on contrastive predictive coding to predict the feature of a clip based on its two neighboring clips. Among the proposed models, the Video Speed baseline showed the best results on the considered downstream tasks.
\begin{figure}[t]
    \centering
    \includegraphics[width=1\textwidth]{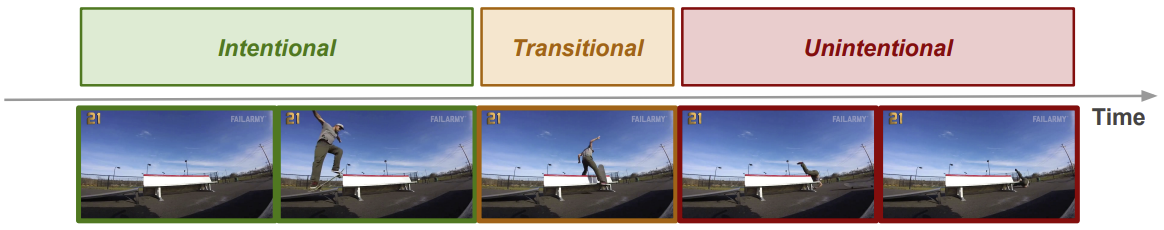}
    \caption{Example of an unintentional action video. Unlike typical videos that are used for action recognition where the labels are fixed throughout the video, the clips labels in unintentional videos change over time. In this example, the skater intentionally jumps at the beginning of the video, but suddenly loses balance and falls down at the end. While all clips are visually similar, we can see the intent degree of the clips changing as the video progresses.}
    \label{fig:unint_label}
\end{figure}

We notice, however, that the proposed methods do not leverage all information present in the unlabeled videos of unintentional actions. 
Firstly, they pay very limited attention to the distinctiveness of features along the temporal axis of the video. For instance, the Video Speed model operates on a temporal context of around one second. For the other two proposed models, the observed local temporal neighborhood is larger, but still limited - three clips of one second are sampled within 0.5 seconds from each other. However, since videos of unintentional actions consist of clips whose labels change throughout the video, the mere distinction of local changes in appearance is insufficient for determining how intentional the portrayed action is.  
For example in Figure~\ref{fig:unint_label}, we can see that all clips look very much alike, however, as the video progresses, the action of the skater turns from intentional to unintentional. This is barely recognizable by looking only at neighboring clips, but the global context of the video reveals that the action does not progress as expected. Secondly, we notice that information contained in the order of sampled clips has not been harnessed to the full extent since the previously proposed Video Sorting model performed only slightly better than the model initialized from scratch. Such relatively low performance, as we show in our experiments, was caused by the specifics of the chosen pre-text task formulation, and not by the lack of information in the clip order. 


In this work, we address the above-mentioned limitations of \cite{Epstein_2020_CVPR}. First, we show that a local neighborhood is insufficient and we propose to sample the clips globally for the entire video. This is required to differentiate between local variations of an intentional action and variations between an intentional and unintentional action. Second, we revisit the poor performance of the Video Sorting approach. While a global sampling improves the learned representation of Video Sorting as well, the permutation-based approach of Video Sorting still struggles in the global setting to distinguish intentional from unintentional actions. We thus propose a pair-wise ordering loss where the features of two randomly sampled clips are concatenated and the network needs to predict which clip occurs first in the video. We demonstrate in the experiments that this loss learns a better representation than the permutation loss where the correct order of three clips needs to be predicted. Finally, we aim to learn a representation that works well for different downstream tasks, namely classification, localization and anticipation of unintentional actions. To this end, we combine a global temporal contrastive loss with a pair-wise video clip order prediction task. 
We show that our proposed approach outperforms all previously proposed self-supervised models for unintentional action prediction and achieves state-of-the-art results on two datasets of unintentional actions: the Oops dataset~\cite{Epstein_2020_CVPR} and LAD2000~\cite{wan2021anomaly}.



\section{Related Work}
\label{related_work}

\subsection{Self-Supervised Learning for Action Recognition}
Self-supervised approaches for video representation learning, as noted by~\cite{Dave2021TCLRTC}, can be generally subdivided into three types: pre-text task-based, contrastive learning-based, and combinations of the two.

\textbf{Video Pretext Task Learning.}
The main idea behind pretext task-based learning is to utilize intrinsic properties of video data that do not require manual annotations for video representation learning. For example, video speed~\cite{Benaim2020SpeedNetLT, Cho2021VPSP, Yao_2020_CVPR} is a commonly used cue for the pretext task formulation. These methods create several versions of the same video with different perceptual speeds by varying their sampling rate and then train the model to recognize the playback speed of the resulting clips. Benaim~\etal ~\cite{Benaim2020SpeedNetLT} formulate this as a binary classification task (normal/sped-up), while other works~\cite{Cho2021VPSP, Yao_2020_CVPR} harness several different sampling rates for multi-class classification.
In addition to speed, Cho~\etal~\cite{Cho2021VPSP} predict the Arrow of Time (AoT) in videos, which was earlier proposed by~\cite{Wei_2018_CVPR}. Recognition of AoT is closely related to another commonly used pre-text task: video order classification~\cite{MisraZH16, Xu_2019_CVPR, Lee_2017_ICCV, FernandoCVPR2017}. Formulations of this task vary. For instance, Misra~\etal~\cite{MisraZH16} predict whether a sequence of frames is in the natural or shuffled order. Lee~\etal~\cite{Lee_2017_ICCV}, on the other hand, predict frame permutations. Xu~\etal~\cite{Xu_2019_CVPR}, like \cite{Lee_2017_ICCV}, predict permutations, but instead of frames they operate on clips. Yet another approach~\cite{FernandoCVPR2017} proposes to recognize a naturally ordered sequence among the shuffled ones.
Other pre-text tasks include, but are not limited to, solving spatio-temporal jigsaw puzzles~\cite{Kim2019SelfSupervisedVR, Ahsan2019VideoJU}, prediction of video rotations~\cite{Jing2018SelfSupervisedSF} and colorization of videos~\cite{Vondrick2018TrackingEB}. 

\textbf{Video Contrastive Learning.}
In recent years, contrastive learning has become popular for self-supervised learning both for image and video domains. Its initial success in the image domain~\cite{Chen2020ASF, He2020MomentumCF} has promoted the emergence of contrastive learning-based methods for video data \cite{Sermanet2017TCN, Qian2021SpatiotemporalCV, Dave2021TCLRTC, Diba_2021_ICCV, Behrmann2021LongSV, HB22}. These methods differ in how they sample positive and negative examples for the training procedure, as well as in the type of the used backbone network (2D or 3D CNN). For many video-based contrastive learning approaches~\cite{Sermanet2017TCN, Qian2021SpatiotemporalCV, Dave2021TCLRTC}, the de facto procedure of constructing examples is by considering clips from the same video as positives and clips from different videos as negatives.
However, enforcing absolute temporal invariance is not beneficial for the resulting features. Therefore, Sermanet~\etal~\cite{Sermanet2017TCN} proposed time contrastive learning, where close frames are considered as positive examples and frames from different parts of the same video as negatives. 
Qian~\etal~\cite{Qian2021SpatiotemporalCV} followed this principle and proposed an extension such that positive examples are sampled from the same video, with the gap between sampled clips drawn randomly from a monotonically decreasing distribution. Dave~\etal~\cite{Dave2021TCLRTC} proposed to train a network with a combination of three loss terms for ensuring different levels of temporal discrimination: instance discrimination, local-local and global-local loss.

\textbf{Combined Learning.}
Some approaches rely on the combination of pre-text task-based and contrastive-based learning. Different methods propose to combine contrastive learning with: speed prediction~\cite{Wang20}, order prediction~\cite{Yao2021SeCoES}, or frame rotation prediction~\cite{knights2020temporally}. Other methods combine several pre-text tasks simultaneously~\cite{Bai2020CanTI, Jenni_2021_ICCV}. Our proposed approach also makes use of the multi-task learning strategy: we combine temporal contrastive learning with order prediction.
However, in contrast to all previous methods that learn representations for action recognition, we learn representations for unintentional action prediction.

\subsection{Self-Supervised Learning for Unintentional Action Recognition}
The task of self-supervised learning for unintentional action prediction has been proposed recently by Epstein~\etal~\cite{Epstein_2020_CVPR}. Previous works on unintentional action prediction or anomaly detection \cite{Wu2020NotOL, Sultani2018RealWorldAD, Sufhakatan2017, Hanson2018BidirectionalCL, Feng_2021_CVPR, Tian_2021_ICCV, Zhong2019GraphCL} do not address representation learning, but focus on predictions based on features pre-extracted from pre-trained networks. Epstein~\etal\cite{Epstein_2020_CVPR, epstein2021learning} instead proposed to learn features specifically for the tasks related to unintentional action prediction. For the unsupervised learning setting, Epstein~\etal~\cite{Epstein_2020_CVPR} proposed three baselines: Video Speed, Video Sorting and Video Context. In the further work, Epstein~\etal\cite{epstein2021learning} have also considered fully-supervised learning, for which they combined learning on labeled examples using the standard cross-entropy loss with solving an unsupervised temporal consistency task. Two further works \cite{zhou2021temporal, xu2022unintentional} addressed the fully-supervised setting for the Oops dataset. While Zhou~\etal~\cite{zhou2021temporal} proposed to model the label distribution to mine reliable annotations for training, Xu~\etal~\cite{xu2022unintentional} proposed a causal inference approach, for which they gathered additional videos with only intentional actions. In this work, we address the challenge of self-supervised representation learning for unintentional action prediction, as defined in~\cite{Epstein_2020_CVPR}.

\section{Method}
\label{method}
We propose to learn representations from videos of unintentional actions using a global temporal contrastive loss and an order prediction loss. In this section, we describe the proposed method in detail.  
We start by formally defining the task of representation learning for unintentional action prediction in Section~\ref{task}. Then, we discuss our temporal contrastive loss in Section~\ref{tmp_contr_loss} and the order prediction loss in Section~\ref{order_loss}.


\subsection{Representation Learning for Unintentionality Detection}
\label{task}
In this work, we deal with unsupervised video representation learning for unintentional action prediction as proposed by~\cite{Epstein_2020_CVPR}. 
Given a set of $V$ unlabeled videos $(v_1, \dots, v_V)$ of unintentional actions, partitioned into $n_i$ short clips $\{x_i^{t}\}_{t=1}^{n_i}$,
the goal is to learn a function $f_i^t=\phi(x_i^t)$ that maps a short clip $x_i^t$ into a feature vector $f_i^t$ in an unsupervised way, such that the resulting features transfer well to downstream tasks related to unintentionality detection. Epstein~\etal\cite{Epstein_2020_CVPR} proposed three such tasks: unintentional action classification, localization and anticipation. The task of classification is to predict labels of individual video clips. The task of localization is to detect a point of transition from an intentional to an unintentional action. Finally, the task of anticipation is to predict the label of a clip located 1.5 seconds into the future. We describe these tasks in more detail in Section~\ref{results}.
For evaluating the learned features on the above mentioned tasks, a small subset of annotated data is provided with labels defined as follows. Initially, each video is annotated with a transitional time point $t_i^a$ at which an action changes from intentional to unintentional. Based on the annotated transitional point $t_i^a$, clips are assigned to one of three classes: intentional, transitional and unintentional. Clip $x_i^t$ is \textit{intentional}, if it ends before the transitional point $t_i^a$; $x_i^t$ is \textit{transitional}, if it starts before $t_i^a$ and ends after it. Otherwise, $x_i^t$ is \textit{unintentional}. An example of a video label annotation is shown in Figure~\ref{fig:unint_label}. 

\begin{figure}[t]
    \centering
    \includegraphics[scale=0.22]{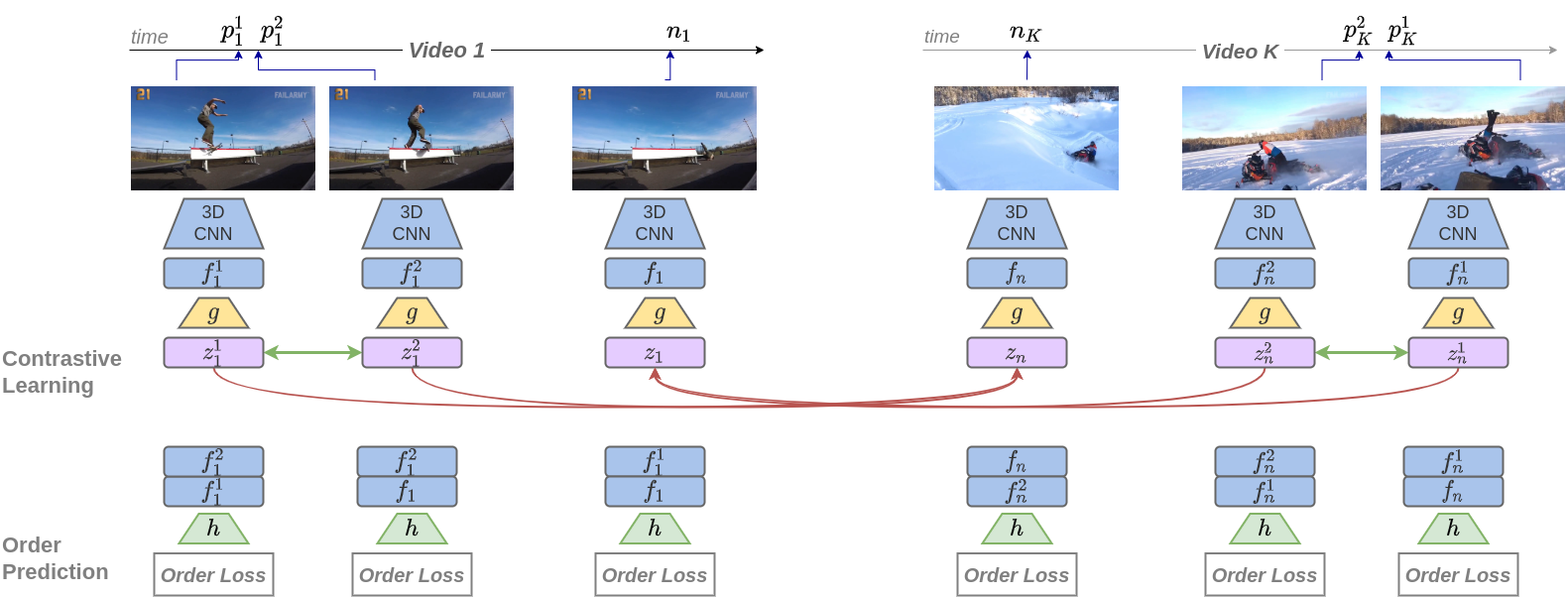}
    \caption{We train our model with a combination of a temporal contrastive loss and an order prediction loss. For each video $v_i$ in the batch of size $K$, we at first randomly sample an anchor clip $p_i^1$ from time step $t_i$. Then, we draw randomly one positive example $p_i^2$ from its immediate neighbors and one negative example $n_i$ from the remaining video clips. Using a backbone 3D-CNN network and a non-linear projection head $g$, we extract features $\{f_i^1, f_i^2, f_i\}_{i=1}^{K}$ and their non-linear projections $\{z_i^1, z_i^2, z_i\}_{i=1}^{K}$ from the sampled clips. We use the $\{z_i^1, z_i^2, z_i\}_{i=1}^{K}$ to compute a temporal contrastive loss (middle row), where positive examples are drawn closer to each other \textit{\color{green}(green arrows)} and are pushed away from the negatives \textit{\color{red}(red arrows)}. Additionally, randomly-ordered pair-wise concatenations of features coming from the same videos $\{f_i^1, f_i^2, f_i\}_{i=1}^{K}$ are used to compute an order prediction loss with a non-linear order prediction head $h$ (bottom row).}
    \label{fig:method}
\end{figure}

\subsection{Temporal Contrastive Loss}
\label{tmp_contr_loss}
As it is motivated in Figure~\ref{fig:unint_label}, we aim to learn a representation that is similar for any clip that contains the same intentional action, but that allows to differentiate if the same action is intentional or unintentional. Since we do not have any labels, we do not know when an action in a video is intentional or unintentional. However, we can assume that sampled neighboring clips are much more likely to have the same label (intentional or unintentional) whereas clips sampled more distant are more likely to have different labels, see Figure~\ref{fig:method}.      
We thus make use of the InfoNCE~\cite{pmlr-v9-gutmann10a} loss function, that encourages representations of two positive views to be closer to each other in the embedding space, while pushing them away from the features of the negative examples. Formally, it is expressed as follows:
\begin{align}
    \label{info_nce_example_loss}
    L_{NCE}(x, y, N) = -\log\frac{q(g(x), g(y))}{q(g(x), g(y)) + \sum_{n \in N}{q(g(x), g(n))}},
\end{align}
where $x, y$ are positive examples, $g$ is a non-linear projection layer, $q(x, y) = \exp(\frac{x \cdot y}{||x|| ||y||} / \tau)$ measures the similarity between samples, which depends on the hyper-parameter $\tau$, and $N$ is the set of negative examples.
For computing the temporal contrastive loss, we obtain positive and negative clips using the following sampling procedure: we sample positive examples from the immediate temporal neighborhood of a given clip while negative examples are sampled from the remaining video regions.  More specifically, for a clip $x_i^t$ from a time point $t$ of video $v_i$, we consider its immediate adjacent clips $\{x_i^{t-1}, x_i^{t+1}\}$ as potential positive examples, whereas clips from the outside regions form a set of potential negative examples, \ie~$\{x_i^{1}, \dots, x_i^{t-3}\} \cup \{x_i^{t+3}, \dots, x_i^{n_i}\}$. Following \cite{Sermanet2017TCN}, clips $\{x_i^{t-2}, x_i^{t+2}\}$ form a margin zone, from which no negative examples are allowed to be sampled.   
Given a mini-batch of $K$ videos $(v_1, \dots, v_K)$, for each video $v_i$ we randomly sample two positive and one negative examples from the set of positives and negatives accordingly. In this way, for each $v_i$ we get a triplet of clips $\{p_i^1, p_i^2, n_i\}_{i=1}^{K}$, where $p_i^1, p_i^2$ is a positive pair and $n_i$ is a negative example. Then, our proposed temporal contrastive loss is formulated as a symmetric InfoNCE loss in the following manner:
\begin{align}\label{eq:temp}
    L_{temp} &= \frac{1}{2K}\sum_{i=1}^{K}(L_{NCE}(p_i^1, p_i^2, N) + L_{NCE}(p_i^2, p_i^1, N)),
\end{align}
where $N = \{n_i\}_{i=1}^{K}$ is the set of all negatives, which is common for all positive pairs. Intuitively, this loss encourages that adjacent clips are close to each other in the embedding space and far apart from the remaining clips. Since we do not restrict the extent of the potential negative regions to a local neighborhood, this loss encourages discrimination between the clips on both fine-grained and more coarse levels. 

\subsection{Order Loss}
\label{order_loss}
While the temporal contrastive loss ensures that neighboring clips are close in the embedding space, it does not ensure that the relative order of the clips gets encoded in the feature vectors. 
However, as indicated by our experiments, order reasoning is key for improving both the classification and the localization of the unintentional moments.
We therefore perform order prediction for each triplet of clips, which are sampled as described in the previous section and illustrated in Figure~\ref{fig:method}. To this end, we do pairwise concatenation of clip features that come from the same videos. For each pair, we randomly select the order in which the representations are concatenated. This results in three concatenated feature vectors for each triplet, \ie~$m=3K$ pairs. Using these pairs, we train a non-linear order prediction layer for recognizing whether representations of the two corresponding clips are concatenated in the correct or reversed order. 
Formally, for two clips $x_i^{t_1}, x_i^{t_2}$ sampled from video $v_i$ at time steps $t_1$ and $t_2$, features $f_i^{t_1}=\phi(x_i^{t_1})$ and
$f_i^{t_2}=\phi(x_i^{t_2})$ 
are concatenated into a single vector 
$f_i^{t_1t_2} = cat(f_i^{t_1}, f_i^{t_2})$ and fed through a non-linear layer $h$ that outputs 
an order prediction vector $\hat{y}_i=h(f_{l_i}^{t_1t_2})$.
The order loss is formulated as the binary cross-entropy loss between predicted labels $\hat{y}_i$ and the correct labels $y_i$: 
\begin{align}
    y_i = \begin{cases} 1, & \text{if } t_1 < t_2 \\
                        0, & \text{otherwise}.
          \end{cases}
\end{align}
Therefore, the final order loss has the following form: 
\begin{align}\label{eq:order}
    L_{ord} &= -\frac{1}{m}\sum_{i=1}^{m} (y_i log(\hat{y_i}) + (1 - y_i)log(1 - \hat{y_i})). 
\end{align}
For training the network, we sum the temporal contrastive loss \eqref{eq:temp} and the order loss \eqref{eq:order}, giving them equal weight:
\begin{align}
    L = L_{temp} + L_{ord}.
\end{align}

\section{Experiments}
\label{experiments}

\subsection{Implementation Details}
We implemented our approach using the Pytorch~\cite{pytorch_2019_NEURIPS} framework. Both for the pre-training stage and downstream tasks, we follow the settings proposed by Epstein~\etal~\cite{Epstein_2020_CVPR} as close as possible for a fair comparison with the previous methods. As a backbone network, we use a randomly initialized ResNet3D-18\cite{Hara2018CanS3}, which receives as input clips of 16 frames from videos sampled at 16 fps, so that one clip contains one second of temporal context.  For the downstream tasks, the pre-trained backbone network is used as a feature extractor and a linear layer is added for the specific task. 
Depending on the task, the backbone network either remains fixed (linear classification and localization) or is finetuned (classification with fine-tuning and anticipation). 
We provide further implementation details in the supplementary material.

\subsection{Datasets}
We conducted our experiments on two datasets with unintentional or anomalous videos: Oops~\cite{Epstein_2020_CVPR} and LAD2000~\cite{wan2021anomaly}. We present the results on the LAD2000 dataset in the supplementary material.
The Oops dataset consists of fail compilation videos from YouTube that have been automatically cut. In total, it has 20723 videos, 16012 train and 4711 test videos, varying from 3 to 30 seconds. Most videos are unlabeled. For evaluating the learned representation on downstreams tasks, the authors provide labeled subsets of the original videos: 6065 training and 4711 test videos. Labeled videos are annotated with a time-step at which the transition from an intentional to an unintentional action happens. 
We further notice that a substantial number of videos in both splits contain several unrelated scenes due to failures of the automatic video cutting. Since our approach is reliant on the temporal coherency of the video data, we do not use such videos during the self-supervised pre-training stage. We therefore perform an automatic detection of videos containing such defects by training a separate model for cut recognition. We provide details about the cut detection network in the supplementary material. We emphasize that this model is trained without manual annotations. For the downstream tasks, we use all videos regardless of their quality for a fair comparison with previous methods. For the sake of completeness, we also include in the supplementary material results of our approach when pre-training is performed on all videos.

\subsection{Results}
\label{results}
In this section, we evaluate the proposed self-supervised learning approach for the different unintentional action recognition tasks.
On the Oops dataset, we report the results on the three tasks proposed by~\cite{Epstein_2020_CVPR}: classification, localization and anticipation. We adopt the protocol of \cite{Epstein_2020_CVPR} and compare our method with other self-supervised approaches, as well as to a network pre-trained on Kinetics~\cite{Kay2017TheKH} with action labels as supervision. Additional comparisons to fully-supervised methods are provided in the supplementary material.

\textbf{Classification.}
The task of classification is to categorize individual one-second-long clips $\{x_i^t\}_{t=1,i=1}^{n_i,n}$ extracted from longer videos containing both intentional and unintentional actions into three categories: intentional, transitional and unintentional.
The clips are uniformly sampled across the videos with increments of 0.25 seconds. To evaluate the proposed self-supervised approach on this task, we fit a linear classifier on top of the features extracted from the backbone pre-trained in a self-supervised way. Following \cite{Epstein_2020_CVPR}, we evaluate the performance of our method for three classification settings. In the first setting \textit{(Linear Classifier, 100\%)}, the backbone network remains fixed and only a linear classification layer is trained on the labeled videos. The second setting \textit{(Linear Classifier, 10\%)} differs from the first one in that only 10\% of the labeled videos are used during training. And finally, in the third setting \textit{(Finetune, 100\%)}, the backbone network is finetuned together with the linear layer using all labeled videos. 
During training, we resample the clips such that the number of clips belonging to the different classes are balanced.

\begin{table}[t]
    \centering
    \begin{tabular}{|C|A|A|B|}
         \hline
         \textbf{Representation} & \multicolumn{2}{c|}{\textbf{Linear Classifier}} & \textbf{Finetune} \\
         \hline
         \% of Labeled Videos & 10\% & 100\% & 100\% \\
         \hline
         \hline
         Kinetics & 52.0 & 53.6 & 64.0 \\
         \hline
         \hline
         Video Speed~\cite{Epstein_2020_CVPR} & 49.9 & 53.4 & 61.6 \\
         Video Context~\cite{Epstein_2020_CVPR} & 47.2 & 50.0 & 60.3 \\
         Video Sorting~\cite{Epstein_2020_CVPR} & 46.5 & 49.8 & 60.2 \\
         \hline
         Scratch~\cite{Epstein_2020_CVPR} & 46.2 & 48.2 & 59.4 \\ 
         \hline
         \textbf{Ours} & \textbf{59.2} & \textbf{60.9} & \textbf{61.9} \\
         \hline
    \end{tabular}
    \caption{Classification accuracy on the Oops dataset.}
    \label{tab:cls_res_oops}
\end{table}

We present the classification results of our approach on the Oops dataset in Table~\ref{tab:cls_res_oops}. All models pre-trained with self-supervision outperform the model trained from scratch for all considered classification settings, which shows the benefit of the self-supervised pre-training. 
For the setting, where only the linear layer is trained, our approach outperforms  by a large margin all previously proposed self-supervised approaches, as well as the model pre-trained on Kinetics with full supervision.
When only $10\%$ of the labeled videos are used for training, our method improves by $9.3\%$ and $7.2\%$ compared to the current state-of-the-art self-supervised approach Video Speed and the representation learned fully-supervised on Kinetics.
When all labeled videos are used, the improvement is $7.5\%$ and $7.3\%$ compared to self-supervised pre-training and pre-training on Kinetics, respectively.
For the fine-tuning case, our model performs on par with the Video Speed \cite{Epstein_2020_CVPR} model. We notice that during the fine-tuning stage almost all self-supervised models perform similarly.

\textbf{Localization.}
The task of localization is to detect the point of transition $t_a$, where the action depicted in the video turns from being intentional to unintentional. To this end, we follow the localization setting proposed by \cite{Epstein_2020_CVPR}. Namely, we use our previously trained linear classifier in a sliding window fashion to estimate the probability of the transitional class for the individual clips of each video. After that, the clip with the highest transitional probability score is considered as the predicted boundary and its center is considered as the predicted transition point $t_a$. This predicted point is deemed to be correct if it has sufficient overlap with the ground-truth transition point. For evaluation, two thresholds for such overlaps are considered: one-quarter second and one second. 

\begin{table}[t]
    \centering
    \begin{tabular}{|W|A|A|}
         \hline
         \textbf{Representation} & \multicolumn{2}{c|}{\textbf{Accuracy within}} \\
         \hline
        \textit{Linear} & 1 sec & 0.25 sec \\
         \hline
         \hline
         Kinetics & 69.2 & 37.8\\
         \hline
         \hline
         Video Speed \cite{Epstein_2020_CVPR} & 65.3 & 36.6  \\
         Video Context \cite{Epstein_2020_CVPR} & 52.0 & 25.3 \\
         Video Sorting \cite{Epstein_2020_CVPR} & 43.3 & 18.3 \\
         \hline
         Scratch~\cite{Epstein_2020_CVPR} & 47.8 & 21.6 \\
         \hline
         \textbf{Ours} & \textbf{70.5} & \textbf{41.3} \\
         \hline
    \end{tabular}
    \caption{Temporal localization accuracy on the Oops dataset.}
    \label{tab:loc_res}
\end{table}
Table~\ref{tab:loc_res} reports evaluation results for the task of transitional point localization. For both considered thresholds our approach outperforms by a substantial margin not only all previously proposed self-supervised methods and the model trained from scratch, but also the model pre-trained in a fully-supervised way on the Kinetics dataset. Compared to the best-performing self-supervised Video Speed model \cite{Epstein_2020_CVPR}, our approach is $5.2\%$ and $4.7\%$ better for one and one-quarter second thresholds, respectively. Whereas for the Kinetics pre-trained model, our method is better by $1.3\%$ and $3.5\%$ for the corresponding thresholds.

\begin{table}[b]
    \centering
    \begin{tabular}{|C|B|B|}
         \hline
         \textbf{Representation} & \multicolumn{2}{c|}{\textbf{Accuracy}} \\
         \hline
         & Reported & Reproduced \\
         \hline
         \hline
         Kinetics & 59.7 & 67.6 \\
         \hline
         \hline
         Video Speed~\cite{Epstein_2020_CVPR} & 56.7 & 61.5  \\
         \hline
         \textbf{Ours} & - & \textbf{62.4} \\
         \hline
    \end{tabular}
    \caption{Anticipation accuracy on the Oops dataset.}
    \label{tab:ant_res}
\end{table}

\textbf{Anticipation.}
The task of anticipation is to predict labels of individual clips 1.5 seconds into the future. As in \cite{Epstein_2020_CVPR}, the task of anticipation is formulated similar to that of the classification with the backbone fine-tuning. However, for this task not all clips can be considered and clip labels are assigned differently. More specifically, for anticipation we can consider only those clips for which there exists a subsequent clip located 1.5 seconds ahead of it. Labels of those future clips are the ones that need to be predicted based on the current clips. We report the anticipation accuracy in Table~\ref{tab:ant_res}. In our experiments, we could not reproduce the results reported in \cite{Epstein_2020_CVPR} for this task, which are shown in the left column (Reported). Thus, we re-evaluated the self-supervised Video Speed model \cite{Epstein_2020_CVPR} using the publicly available weights and the model pre-trained on Kinetics (Reproduced). Our method outperforms the previously proposed self-supervised approach Video Speed by $0.9\%$.  

\subsection{Ablation study}
\label{ablation}
In this section, we present a set of ablation experiments to study the impact of the different aspects of our approach. Firstly, we provide an analysis of the individual components of our proposed loss. Secondly, we study the impact of the temporal extent and the order loss formulation on the quality of the learned  representations and thereby uncover the limitations of the previously proposed approaches.
All ablation studies are performed on the Oops dataset on both the classification and localization tasks by training a linear classifier on top of the learned representations.

\textbf{Effect of Loss Components.}
The loss function for our model consists of two components: temporal contrastive loss and order prediction loss. To analyze the impact of the individual loss terms, we train two separate models: one with temporal contrastive loss $(L_{temp})$, and another one with order prediction loss $(L_{ord})$. The results for these models are shown in Table~\ref{tab:abl_loss}. We can observe for both downstream tasks that using both loss terms substantially outperforms the models trained with only one of the two loss terms. This supports our hypothesis that temporal cues and relative order information are complementary and both are crucial for recognizing the intentionality of human actions.
We can further see that while order-based and contrastive-based models show similar performance for localization, the order-based approach achieves significantly better results on the task of classification. 

\begin{table}[t]
    \centering
    \begin{tabular}{|C|A|A|B|}
    \hline
    \multicolumn{4}{|c|}{\textbf{Linear classifier}} \\
    \hline
    & \multicolumn{2}{c|}{\textbf{Localization}} & \textbf{Classification} \\
    \hline
    \textbf{Loss} & 1 sec & 0.25 sec & All labels  \\
    \hline
    $L_{temp}$ & 65.7 & 37.1 & 52.3 \\
    \hline
    $L_{ord}$ & 66.2 & 36.0 & 60.0 \\
    \hline
    $L_{temp} + L_{ord}$ & \textbf{70.5} & \textbf{41.3} & \textbf{60.9} \\
    \hline
    \end{tabular}
    \caption{Impact of the loss components on the Oops dataset.}
    \label{tab:abl_loss}
\end{table}

\textbf{Effect of Loss Formulation.}
For computing the temporal contrastive loss and the order prediction loss, we sample three clips per video: two positives and one negative. To analyze the impact of the temporal variety of the samples on the quality of the resulting features, we perform a set of experiments. 
Namely, we train the models with the corresponding loss terms but we change the temporal scope from which the negative clips are sampled. We refer to the original sampling setting as \textit{Global}, while \textit{Local} denotes that the sampling scope has been limited. For \textit{Local}, potential negative examples are confined to clips that are at least 3 and at most 5 steps away from the anchor clip $x_i^t$, \ie~$\{x_i^{t-5}, \dots, x_i^{t-3}\} \cup \{x_i^{t+3}, \dots, x_i^{t+5}\}$.
In addition to this, we also analyze how the formulation of the order prediction task affects the resulting features. More precisely, we experiment with the formulation proposed by Epstein~\etal~\cite{Epstein_2020_CVPR}. That is, instead of predicting the pairwise order of clips (\textit{Pair Order}), we predict their random permutation (\textit{Permutation}).

Table~\ref{tab:abl_formulation} shows the experimental results. Firstly, we observe that limiting the temporal variety of the considered examples generally leads to a weaker performance on the downstream tasks. This confirms our hypothesis that both long-range and short-range temporal discrimination are crucial for recognizing the degree of intent in a video. Therefore loss functions should not be confined to local neighborhoods. Secondly, we can see that models trained for permutation prediction show significantly lower results than those trained with pair-wise order prediction, especially when only clips from the confined local neighborhood are considered, which is a setting similar to what is proposed by Epstein~\etal\cite{Epstein_2020_CVPR}. This shows that our formulation of order prediction is better suited to learn representations for unintentional action prediction, since it harnesses the information available in the relative clip order more efficiently. 
\begin{table}[t]
    \centering
    \begin{tabular}{|B|B|A|A|B|}
    \hline
    \multicolumn{5}{|c|}{\textbf{Linear classifier}} \\
    \hline
    \multicolumn{2}{|c|}{} & \multicolumn{2}{c|}{\textbf{Localization}} & \textbf{Classification} \\
    \hline
    \multicolumn{2}{|c|}{\textbf{Loss}} & 1 sec & 0.25 sec & All labels  \\
    \hline
    Temp. ext. & Loss Type &  & &   \\
    \hline
    \hline
    
    \multicolumn{5}{|c|}{\textbf{Temporal Contrastive} ($L_{temp}$)} \\
    \hline
    \hline
    \textit{Global} & - &  65.7 & 37.1  & 52.3 \\
    \hline
    \textit{Local} & - & 64.1 & 35.5 & 50.1 \\
    \hline
    \hline
    
    \multicolumn{5}{|c|}{\textbf{Order} ($L_{ord}$)} \\
    \hline
    \textit{Global} & \textit{Pair Order} & 66.2 & 36.0 & 60.0 \\
    \hline
    \textit{Local} & \textit{Pair Order} & 66.6 & 36.9 &  57.4 \\
    \hline
    \textit{Global} & \textit{Permutation} & 59.7 & 32.1 & 47.2 \\
    \hline
    \textit{Local} & \textit{Permutation} & 46.5 & 20.1 & 46.6  \\
   
    \hline
    \hline
    \multicolumn{5}{|c|}{\textbf{Temporal Contrastive $+$ Order} ($L_{temp} + L_{ord}$)} \\
    \hline
    \textit{Global} & \textit{Pair Order} & 70.5 & \textbf{41.3} & \textbf{60.9} \\
    \hline
    \textit{Local} & \textit{Pair Order} & \textbf{70.7 }  & 40.8 & 57.2  \\
    \hline
    \end{tabular}
    \caption{Impact of the loss formulation on the Oops dataset.}
    \label{tab:abl_formulation}
\end{table}


\section{Discussion}
In this work, we have presented a self-supervised approach for the task of unintentional action prediction. To address the shortcomings of the previously proposed approaches, we proposed to train a feature extraction network using a two-component loss, consisting of a temporal contrastive loss and an order prediction loss. While the temporal contrastive loss is responsible for making representations distinct, the prediction of the clip order allows to learn the relative positioning. We evaluated our approach on one video dataset with unintentional actions and we provided additional results for anomaly detection in the supplementary material.   
While our work shows a large improvement over the previously proposed methods, we recognize that the current evaluation setting has a number of limitations. 
More specifically, the Oops dataset that we used for pre-training our model consists only of videos that have been collected from fail compilation videos. These videos are biased towards very specific actions and scenes, and the temporal location of unintentional actions is biased as well. While we show in the supplementary material that learning representations on such videos can be beneficial for anomaly detection as well, it needs to be investigated in future work how well the proposed approaches perform on videos with a more general definition of unintentional actions. 

\paragraph{Acknowledgments.}
The work has been funded by the Deutsche Forschungsgemeinschaft (DFG, German Research Foundation) – SFB 1502/1–2022 - Projektnummer: 450058266 and GA 1927/4-2 (FOR 2535 Anticipating Human Behavior).
\bibliographystyle{splncs04}
\bibliography{egbib}

\end{document}